\pdfoutput=1
\documentclass{article}

\PassOptionsToPackage{numbers, compress}{natbib}

\usepackage[preprint]{neurips_2023}

\usepackage[utf8]{inputenc} 
\usepackage[T1]{fontenc}    
\usepackage{hyperref}       
\usepackage{url}            
\usepackage{booktabs}       
\usepackage{amsfonts}       
\usepackage{nicefrac}       
\usepackage{microtype}      
\usepackage{amsmath}
\usepackage{natbib}
\usepackage{adjustbox}
\usepackage{algorithmic}
\usepackage{graphicx}
\usepackage{textcomp}
\usepackage[table]{xcolor}
\usepackage{multirow}
\usepackage{changepage,threeparttable}
\usepackage{makecell}
\usepackage{xspace}
\usepackage{enumitem}
\usepackage{wrapfig}
\usepackage{tikz}

\usepackage[colorinlistoftodos]{todonotes}

\usepackage{comment}

\usepackage{caption}
\usepackage{subcaption}
\usepackage{tikz-cd}

\usepackage{amsmath,amsfonts,amssymb,amsthm,bm}

\def\eqref#1{equation~\ref{#1}}

\def\1{\bm{1}}

\def\eps{{\epsilon}}

\DeclareMathAlphabet{\mathsfit}{\encodingdefault}{\sfdefault}{m}{sl}
\SetMathAlphabet{\mathsfit}{bold}{\encodingdefault}{\sfdefault}{bx}{n}

\def\gB{{\mathcal{B}}}

\usepackage{cellspace, multirow, booktabs}
\newtheorem{definition}{Definition}

\newtheorem{theorem}{Theorem}

\usepackage{quiver}

\usepackage{stmaryrd}

\usepackage{graphicx}
\usepackage{tikz-cd}

\newcommand{\style}[1]{{\emph{#1}}}  

\newcommand{\face}{{\trianglelefteqslant}}

\newcommand{\Fc}{\mathcal{F}}

\DeclareMathOperator{\id}{id}

\def\@lbibitem[#1]#2{\item[{[#1]}\hfill]\if@filesw
      {\let\protect\noexpand
       \immediate
       \write\@auxout{\string\bibcite{#2}{#1}}}\fi\ignorespaces}

\definecolor{gold}{RGB}{221, 196, 65}
\definecolor{silver}{RGB}{215, 215, 215}
\definecolor{bronze}{RGB}{205, 127, 50}
\newcommand{\tikzcircle}[2][red,fill=red]{\tikz[baseline=-0.7ex]\draw[#1,radius=#2] (0,0) circle ;}%

\newcommand{\peptides}{\texttt{Peptides}\xspace}
\newcommand{\pepfunc}{\texttt{Peptides-func}\xspace}
\newcommand{\pepstruct}{\texttt{Peptides-struct}\xspace}

\definecolor{lightgray}{gray}{0.95}
\definecolor{Gray}{gray}{0.85}
\definecolor{LightCyan}{rgb}{0.88,1,1}
\definecolor{LightPink}{HTML}{FCE1EF}
\definecolor{LightGreen}{HTML}{EEF7E1}
\newcolumntype{A}{>{\columncolor{white}}c}
\newcolumntype{B}{>{\columncolor{LightGreen}}c}
\newcolumntype{C}{>{\columncolor{LightPink}}c}

\definecolor{purpleheart}{rgb}{0.41, 0.21, 0.61}

\definecolor{dark2green}{rgb}{0.1, 0.65, 0.3}
\definecolor{dark2orange}{rgb}{0.9, 0.4, 0.}
\definecolor{dark2purple}{rgb}{0.4, 0.4, 0.8}
\newcommand{\first}[1]{\colorbox{gold}{\textbf{#1}}}
\newcommand{\second}[1]{\colorbox{silver}{#1}}
\newcommand{\third}[1]{\colorbox{bronze}{{#1}}}

\definecolor{lightgreen}{RGB}{168,207,147}
\definecolor{lightblue}{RGB}{34,118,180}
\definecolor{lightyellow}{RGB}{255,226,149}

\usepackage{tcolorbox}

\makeatletter

\setbox0\hbox{$\xdef\scriptratio{\strip@pt\dimexpr
    \numexpr(\sf@size*65536)/\f@size sp}$}

\newcommand{\scriptveryshortarrow}[1][3pt]{{%
    \hbox{\rule[\scriptratio\dimexpr\fontdimen22\textfont2-.2pt\relax]
               {\scriptratio\dimexpr#1\relax}{\scriptratio\dimexpr.4pt\relax}}%
   \mkern-4mu\hbox{\let\f@size\sf@size\usefont{U}{lasy}{m}{n}\symbol{41}}}}

\makeatother

\title{CIN++: Enhancing Topological Message Passing}

\author{%
  Lorenzo ~Giusti\thanks{The work was performed while the author was in Cambridge.} \\
  Sapienza University\\
  \texttt{lorenzo.giusti@uniroma1.it} \\
  \And
    Teodora ~Reu\\
    University of Cambridge\\
    \texttt{tr500@cam.ac.uk} \\
  \And
    Francesco ~Ceccarelli\\
    University of Cambridge\\
    \texttt{fc485@cam.ac.uk} \\
  \And
    Cristian ~Bodnar\\
    Microsoft Research AI4Science\\
    \texttt{cbodnar@microsoft.com} \\
  \And
    Pietro ~Li\`{o}\\
    University of Cambridge\\
    \texttt{pl219@cam.ac.uk} \\
}

\begin{document}

\maketitle

\begin{abstract}
Graph Neural Networks (GNNs) have demonstrated remarkable success in learning from graph-structured data. However, they face significant limitations in expressive power, struggling with long-range interactions and lacking a principled approach to modeling higher-order structures and group interactions. Cellular Isomorphism Networks (CINs) recently addressed most of these challenges with a message passing scheme based on cell complexes. Despite their advantages, CINs make use only of boundary and upper messages which do not consider a direct interaction between the rings present in the underlying complex. Accounting for these interactions might be crucial for learning representations of many real-world complex phenomena such as the dynamics of supramolecular assemblies, neural activity within the brain, and gene regulation processes. In this work, we propose CIN++, an enhancement of the topological message passing scheme introduced in CINs. Our message passing scheme accounts for the aforementioned limitations by letting the cells to receive also lower messages within each layer. By providing a more comprehensive representation of higher-order and long-range interactions, our enhanced topological message passing \footnote{The code implementation can be found at: \url{https://github.com/twitter-research/cwn}} scheme achieves state-of-the-art results on large-scale and long-range chemistry benchmarks.
\end{abstract}

\section{Introduction}

Graph Neural Networks (GNNs) find applications in a plethora of fields, like computational chemistry~\cite{gilmer2017neural}, social networks~\cite{fan2019graph} and physics simulations~\cite{sanchez2020learning}. Since their introduction~\cite{gori2005new, scarselli2008graph}, GNNs have shown remarkable results in learning tasks when data are defined over a graph domain, where the flexibility of neural networks is coupled with prior knowledge about data relationships, expressed in terms of the underlying topology~\cite{battaglia2018relational}. The idea behind GNNs is learning representations of node features using local aggregation, where the neighbourhood is formally represented by the underlying graph, which can be seen as a simple instance of a \emph{topological space}, able to capture {\it pairwise} interactions through the presence of an edge between any pair of directly interacting nodes. By leveraging this simple but powerful idea, outstanding performance has been achieved in many traditional tasks such as classification for nodes or entire graphs~\cite{kipf2016semi} or link prediction~\cite{zhang2018link} as well as more specialized ones such as \emph{protein folding}~\cite{jumper2021highly} and \emph{algorithmic reasoning}~\cite{velivckovic2021neural}.  While Graph Neural Networks (GNNs) have advanced the modelling of pairwise interactions on graph-structured data, their inability to accurately capture long-range and group interactions, along with their struggles to manage higher-order structures, are significant shortcomings. These limitations critically restrict the application of GNNs in understanding real-world complex systems.

To cope with these limitations, a major performance boost to GNNs algorithms has been offered by considering more complex topological spaces such as simplicial complexes~\cite{ebli2020simplicial} or cell complexes~\cite{hajij2020cell}, introduced to handle tasks for data that are naturally defined on higher-order elements. Then, these ideas were combined with provably powerful message-passing schemes on simplicial~\cite{bodnar2021weisfeiler} and cell complexes~\cite{bodnarcwnet} achieving remarkable results. Nonetheless, the aforementioned models are unable to discover consistently long-range and group interactions, which play a crucial role in many practical applications such as network neuroscience~\cite{giusti2016two}, physics of complex systems~\cite{lambiotte2019networks} or gene regulatory networks~\cite{sever2015signal}, where some reactions occur only when a group of more than two entities interact.

\textbf{Contribution} In this work, we leverage the advantages of complex topological spaces to introduce a novel message-passing scheme on cell complexes. Motivated by the fact that cell complexes provide a natural framework to represent higher-dimensional structures and topological features that are inherent in the realm of chemistry, throughout this work, we will mostly focus on this domain. In particular, we enhanced the Topological Message Passing scheme defined in~\cite{bodnarcwnet} by including messages that flow within the lower neighbourhood of the underlying cell complex.  These are messages exchanged between edges that share a common vertex and between rings that are glued through an edge to better capture group interactions and escape potential bottlenecks. In the experimental section, we show that with respect to other models, this representation allows for a more natural and comprehensive understanding of chemical systems and their properties, resulting in state-of-the-art performance in both a large-scale molecular benchmark (ZINC) and a long-range graph benchmark (Peptides). We see that the ability of our model to understand higher-dimensional structures and topological features could have an immediate and significant impact in the areas of computational chemistry and drug discovery.


\section{Background}\label{sec:background}
\begin{figure}[t]
    \centering
    \vspace{-40pt}
    \includegraphics[width=.9\textwidth]{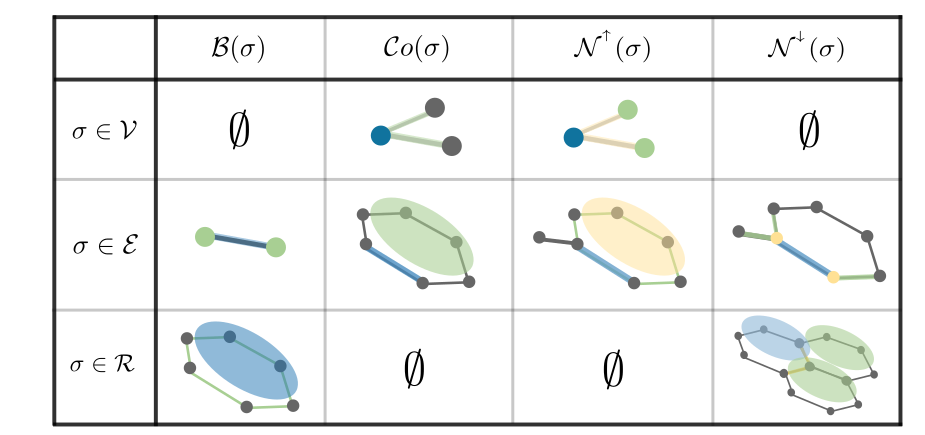}
    \caption{Visual representation of adjacencies within cell complexes. The reference cell, $\sigma$, is showcased in \textcolor{lightblue}{blue}, with adjacent cells $\tau$ highlighted in \textcolor{lightgreen}{green}. Any intermediary cells, $\delta$, facilitating connectivity are depicted in \textcolor{lightyellow}{yellow}.}
    \label{fig:adjs}
\end{figure}

In this section, we recall the basics of regular cell complexes. These are topological spaces that enable efficient representation of high-order interaction systems, generalizing graphs and simplicial complexes. In particular, we first introduce the  definition of a regular cell complex and then we recall a few additional properties enabling the representation of cell complexes via boundary operators.

\begin{definition}[Regular Cell Complex]\label{def:cell_complex}
\cite{hansen2019toward} \textit{A {\it regular cell complex} is a topological space $\mathcal{C}$ together with a partition $\{\mathcal{C}_{\sigma}\}_{\sigma \in \mathcal{P}_{\mathcal{C}}}$ of subspaces $\mathcal{C}_{\sigma}$ of $\mathcal{C}$ called $\mathbf{cells}$, where $\mathcal{P}_{\mathcal{C}}$ is the indexing set of $\mathcal{C}$, such that}

\begin{enumerate}
    \item For each $\sigma$ $\in$  $\mathcal{C}$, every sufficient small neighbourhood of $\sigma$ intersects finitely many $\mathcal{C}_{\sigma}$;  
    \item \label{it:closure} For all $\mathcal{\tau}$,$\mathcal{\sigma}$ we have that $\mathcal{C}_{\tau}$ $\cap$ $\overline{\mathcal{C}}_{\sigma}$ $\neq$ $\varnothing$ iff $\mathcal{C}_{\tau}$ $\subseteq$ $\overline{\mathcal{C}}_{\sigma}$, where $\overline{\mathcal{C}}_{\sigma}$ is the closure of the cell;
    \item \label{it:homeomo} Every $\mathcal{C}_{\sigma}$ is homeomorphic to $\mathbb{R}^{k}$ for some $k$;
    \item For every $\sigma$ $\in$ $\mathcal{P}_{\mathcal{C}}$ there is a homeomorphism $\phi$ of a closed ball in $\mathbb{R}^{k}$ to $\overline{\mathcal{C}}_{\sigma}$ such that the restriction of $\phi$ to the interior of the ball is a homeomorphism onto $\mathcal{C}_{\sigma}$.
\end{enumerate}
\end{definition}

Condition \ref{it:closure} implies that the indexing set $\mathcal{P}_{\mathcal{C}}$ has a poset structure, given by $\tau$ $\leq$ $\sigma$ iff $\mathcal{C}_{\tau}$ $\subseteq$ $\overline{\mathcal{C}_\sigma}$. This is known as the face poset of $\mathcal{C}$. The regularity condition (4) implies that all topological information about $\mathcal{C}$ is encoded in the poset structure of $\mathcal{P}_{\mathcal{C}}$. Then, a regular cell complex can be identified with its face poset. For this reason, from now on we will indicate the cell $\mathcal{C}_{\sigma}$ with its corresponding face poset element $\sigma$ which dimension $\dim({\sigma})$ is equal to the dimension of the space homeomorphic to $\mathcal{C}_{\sigma}$. In this study, we focus on cell complexes with cells of maximum dimension equal to 2. 

\begin{wrapfigure}[15]{r}{0.3\textwidth}
    \begin{subfigure}[t!]{1.0\linewidth}
        \centering
        \vspace{-20pt}
        \includegraphics[width=1.0\textwidth]{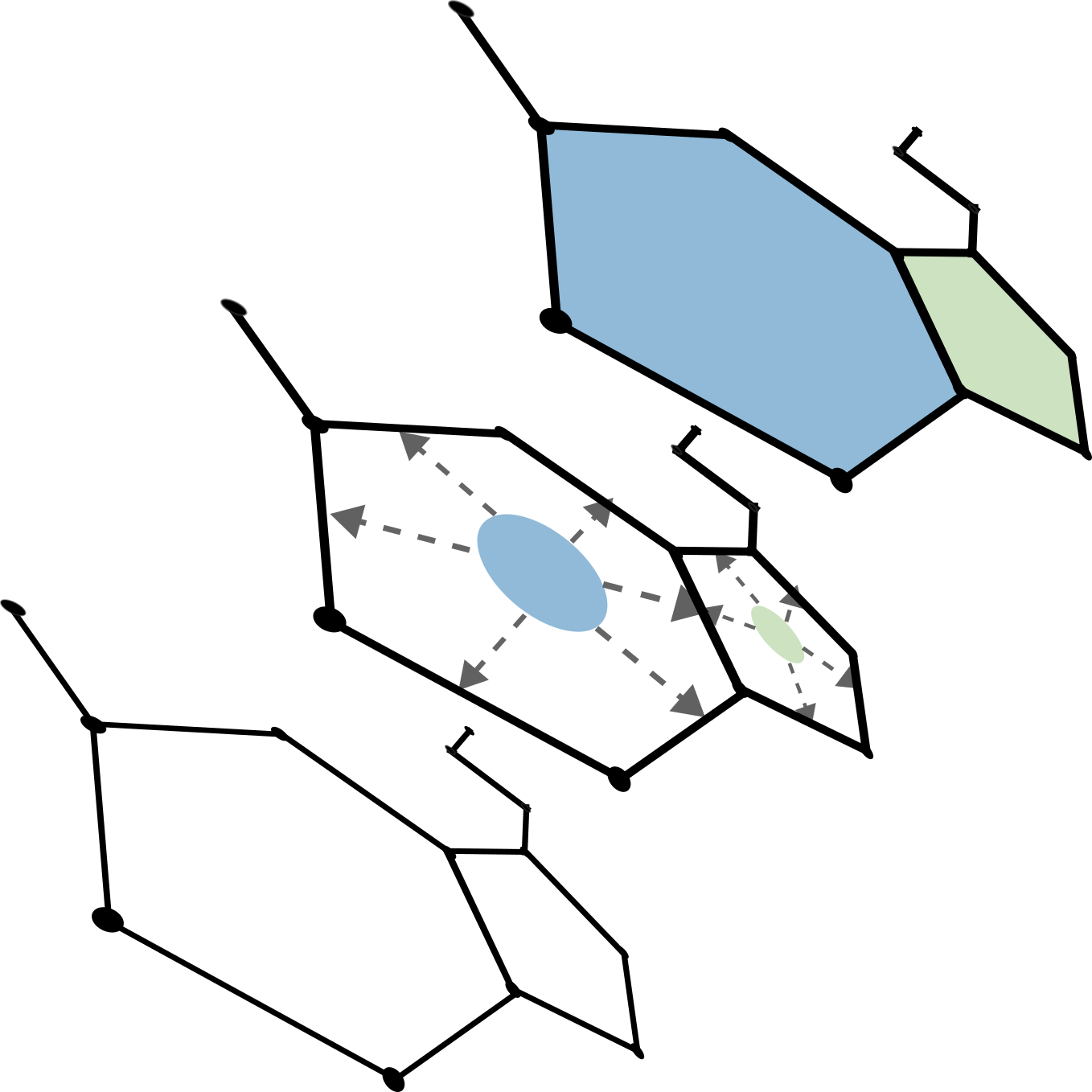}
    \end{subfigure}
     \caption{Cellular lifting process. Given an input graph $\mathcal{G}$, we attach closed two-dimensional rings to the boundary of the induced cycles of $\mathcal{G}$. The result is a 2D regular cell complex $\mathcal{C}$.}
    \label{fig:sl}
    \vspace{-25pt}
\end{wrapfigure}

In this context, a graph $\mathcal{G} = (\mathcal{V}, \mathcal{E})$ can be viewed as a particular case of a regular cell complex $\mathcal{C}$. Specifically, a graph is a cell complex  where the set of 2-cells is the empty set. In this context, the vertices of the graph correspond to the 0-cells in $\mathcal{C}$, while the edges of the graph are then represented by its 1-cells, connecting pairs of vertices. Throughout this work, we will consider regular cell complexes $\mathcal{C}$ built using {\em skeleton-preserving cellular lifting maps}~\cite{bodnarcwnet} from an input graph $\mathcal{G}$. A pictorial example of this operation is provided in Fig.~\ref{fig:sl}, where filled rings are attached to closed paths of edges having no internal chords.

\
\

\begin{definition}[Boundary Relation]\label{def:boundary_rel}
    Given two cells $\sigma, \tau \in \mathcal{C}$. We have the boundary relation  $\sigma$ $\face$ $\tau$ \emph{iff} $\dim({\sigma}) \leq \dim({\tau})$ and $\nexists \delta \in \mathcal{C} :  \sigma \face \delta \face \tau$.
\end{definition}

We can leverage the previous definitions to characterize the different types of neighbourhoods present in cell complexes.

\paragraph{Boundary Neighbourhood} For a cell $\sigma$, the boundary is a set $\mathcal{B}(\sigma) = \{ \tau \, \vert \,  \tau \face \sigma \}$ composed by the lower-dimensional cells that respect Definition~\ref{def:boundary_rel}. In the first column of Fig.~\ref{fig:adjs} we depicted a glossary of the boundary neighbourhoods of a regular cell complex $\mathcal{C}$. In particular, a vertex does not have a boundary neighbourhood, an edge has a boundary composed of the nodes attached to its endpoints, while the boundary cells of a ring are the edges that enclose the ring itself.

\paragraph{Co-Boundary Neighbourhood} For a cell $\sigma$, the co-boundary neighbourhood is a set $\mathcal{C}o(\sigma) = \{\tau \, \vert \, \sigma \face \tau \}$ of higher-dimensional cells with $\sigma$ on their boundary. For a node, its co-boundary is composed of the edges that have that node as an endpoint. For an edge, it is the set of rings that have that edge as one of their sides. In our case, rings do not have a co-boundary neighbourhood. We show a pictorial example of the various co-boundary neighbourhoods in the second column of Fig.~\ref{fig:adjs}.

\paragraph{Upper Neighbourhood} These are the cells of the same dimension as $\sigma$ that are on the boundary of the same higher-dimensional cell as $\sigma : \mathcal{N}^{{}^{\uparrow}}{(\sigma)} = \{ \tau \, \vert \, \exists \delta : \sigma \face \delta \wedge \tau \face \delta\}$. For instance, as shown in the third column of Fig.~\ref{fig:adjs}, the upper adjacent cells of a vertex $v_i$ are the vertices connected to $v_i$ via an edge (i.e., the canonical graph adjacency). The upper adjacent cells of an edge $e_i$ are the edges that surround the rings for which $e_i$ is a boundary element. However, in a 2-Complex, the rings do not have upper adjacent cells.

\paragraph{Lower Neighbourhood} These are the cells of the same dimension as $\sigma$ that share a lower dimensional cell on their boundary: $\mathcal{N}^{{}^{\downarrow}}{(\sigma)} = \{ \tau \, \vert \, \exists \delta : \delta \face \sigma \wedge \delta \face \tau\}$. For instance, as shown in the fourth column of Fig.~\ref{fig:adjs}, the lower adjacent cells of an edge $e_i$ are the edges that share a common vertex with $e_i$ and the lower adjacent cells of a ring $r_i$ are the rings that have a common edge on their boundary. In any case, the vertices of a regular cell complex do not have a lower neighbourhood. 

After defining the structural and neighbourhood elements of a cell complex, we now address how to represent signals over it. A cell signal is defined as a mapping from the set of all the cells  contained in the complex to a multi-dimensional feature vector~\cite{giusti2022cell}.
 
\begin{definition}[Signals Over Cell Complexes]
    Let $\mathcal{C}$ be a cell-complex and let $\mathcal{P}_{\mathcal{C}}$ be its indexing set. A cell signal is defined as a map $h_{\sigma} : \mathcal{P}_{\mathcal{C}}\rightarrow \mathbb{R}^d$ that assigns a $d$-dimensional feature vector to each cell $\sigma$ of the complex. 
\end{definition}

\section{Supramolecular Chemistry}

Supramolecular chemistry~\cite{steed2022supramolecular}, often referred to as the {\em "chemistry beyond the molecule"}, explores the intricacies of molecules connected through various weak bonds of differing strengths. These spontaneous secondary interactions include hydrogen bonding, dipole-dipole, charge transfer, van der Waals, and $\pi-\pi$ stacking interactions. The presence of numerous $\pi-\pi$ stacking interactions is particularly significant in this context, as the overall system can be seen as a regular cell complex structure. Supramolecular assemblies often exhibit complex chemical architectures and high-order self-assembly, giving rise to molecular machines~\cite{feringa2011molecular}, gas absorption~\cite{millward2005metal}, high-tech molecular sensing systems~\cite{allendorf2009luminescent}, nanoreactors~\cite{mattia2015supramolecular}, chemical catalysis~\cite{lee2009metal} and drug delivery systems~\cite{webber2017drug}. Intriguingly, molecular shape serves as a design principle, thanks to the self-assembly~\cite{whitesides2002self} and self-healing~\cite{white2001autonomic} properties of supramolecules. 

\subsection{Long-Range Interactions}

\begin{wrapfigure}[22]{r}{0.25\textwidth}
    \begin{subfigure}[t!]{1.0\linewidth}
        \centering
        \includegraphics[width=1.0\textwidth]{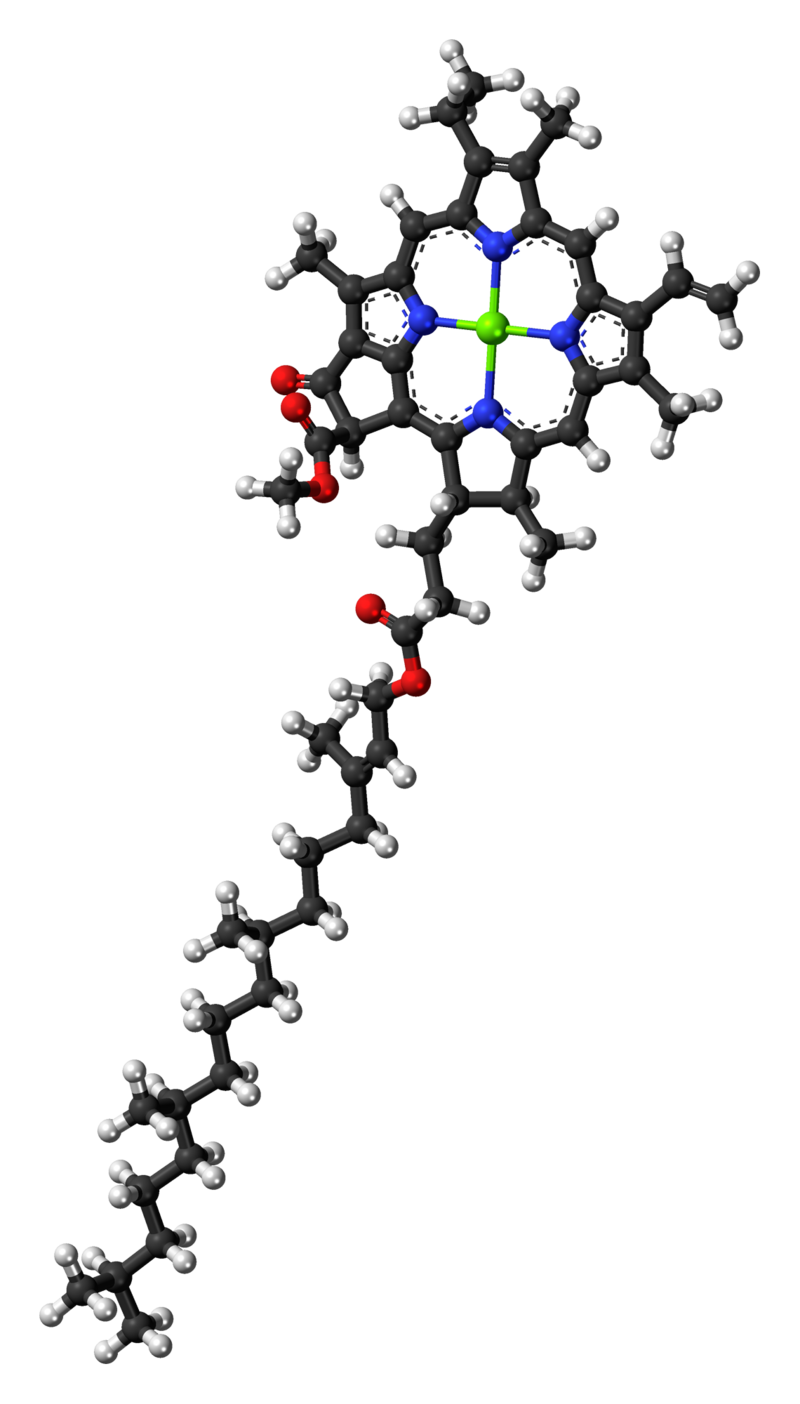}
    \end{subfigure}
     \caption{Structure of Chlorophyll-a, the most common molecule in photosynthetic organisms.}
    \label{fig:clorophylla}
\end{wrapfigure}
\
\

Long-range interactions play a key role in supramolecular chemistry. They can be intended as the dependency of certain molecular properties on elements that are {\em "far away"} within a chemical system~\cite{gray2005long}. Of particular interest in this context are the long-range interactions that arise in oxygenic photosynthesis. This is the process by which light energy is converted into chemical energy in the form of glucose or other sugars~\cite{barber2009photosynthetic}. This process is mediated by Chlorophyll-a (Fig.\ref{fig:clorophylla}), a cyclic tetrapyrrole molecule. Through its extensive conjugated $\pi$-system, Chlorophyll-a represents the basic building block of a photosystem. During photosynthesis, when a photon strikes a molecule of Chlorophyll-a, it excites an electron to a higher energy state. The energy produced is transferred from molecule to molecule within the light-harvesting complex via resonance energy transfer. Throughout the whole process, the energy transfer process is materialised as a quantum-coherent phenomenon ~\cite{engel2007evidence}, that is where long-range interactions become crucial. Being able to capture them could lead to a positive impact in the development of efficient artificial photosynthetic systems~\cite{gust2001mimicking} and enhance solar energy technologies~\cite{green2021solar}.

\paragraph{On Oversquashing in Molecular Graphs}  
To capture interactions in molecular graphs, in the last years we have seen {\em Message Passing Neural Networks} (MPNNs)~\cite{gilmer2017neural} taken as a reference model with remarkable results. These are a class of Graph Neural Networks (GNNs) that update the spatial representation of a node $u$ with layers of the form: 
\begin{equation}\label{eq:mpnn}
    \mathbf{h}_{u}^{l+1} = \text{U}\Bigl(\mathbf{h}_{u}^{l},\underset{v \in \mathcal{N}(u)}{\text{AGG}}\Bigl(\mathbf{h}_{v}^l\Bigr)\Bigr),
\end{equation}

where $\text{U}$ is a function that {\em updates} the node's current features with messages from its neighbours and $\text{AGG}$ is a {\em permutation-invariant aggregation function}. When it is required to aggregate information between nodes located in remote parts of the graph, MPNNs as in Eq.~\ref{eq:mpnn} are susceptible to bottlenecks. These bottlenecks are manifested as an exponentially increasing amount of information constrained into vectors with a fixed representation. This is known in the literature as {\em Oversquashing}~\cite{alon2020bottleneck, di2023over}, a phenomenon that leads to sub-optimal performance when the prediction task is highly reliant on long-range interactions. Oversquashing arises in MPNNs  because the propagation of information happens between nodes that are connected through edges, which induces a computational graph directly mirroring the input graph structure.

Message passing schemes on complex topological spaces or {\em Topological Message Passing}~\cite{bodnar2021topological, bodnarcwnet, hajij2020cell} mitigate this issue by not just considering nodes (0-cells) and edges (1-cells), but also involving higher-dimensional elements such as rings (2-cells). With a richer topological structure,  the messages can be propagated through these higher-dimensional cells, effectively providing shortcuts or additional routes for information flow. With this construction, the underlying computational graph is no longer coupled with the input graph structure.

\begin{wrapfigure}[14]{r}{0.25\textwidth}
    \begin{subfigure}[t!]{1.0\linewidth}
        \centering
        \vspace{-20pt}
        \includegraphics[width=1.0\textwidth]{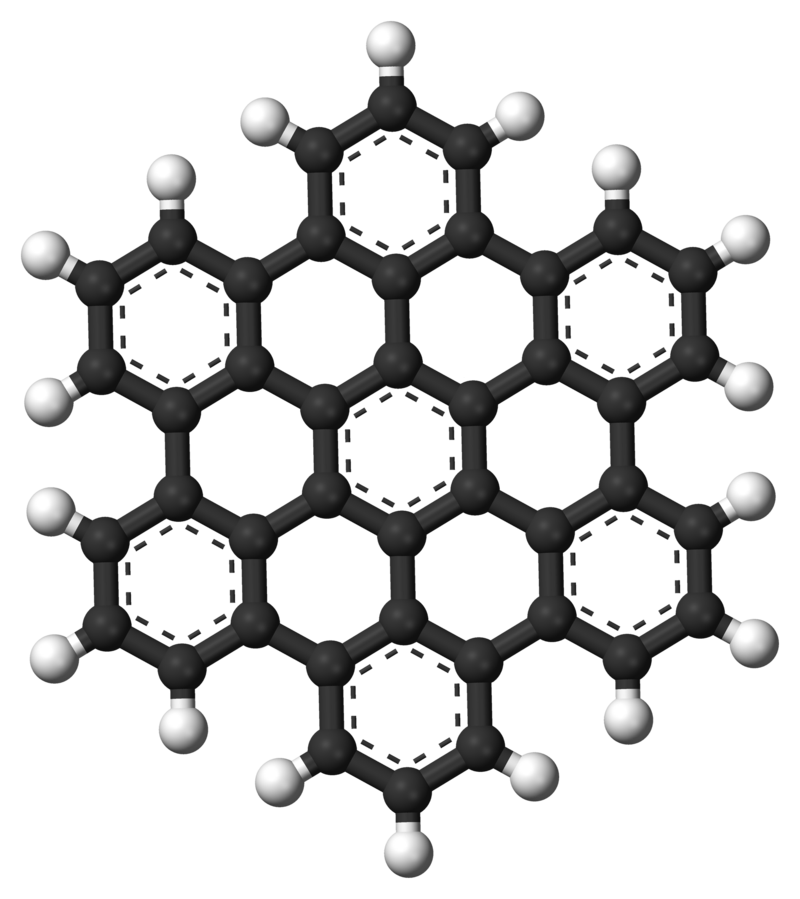}
    \end{subfigure}
     \caption{Representation of hexabenzocoronene, a polycyclic aromatic hydrocarbon.}
    \label{fig:pah}
\end{wrapfigure}

\subsection{Group Interactions} As long-range interactions, group interactions play a fundamental role in chemical and biological processes. One example is the case of aromatic stacking. Aromatic stacking refers to the non-covalent interactions between aromatic rings, such as those found in the amino acid tryptophan or the nucleotide bases of DNA~\cite{hunter1990nature}. These interactions are essential in various biological processes, {\em including protein folding, DNA/RNA structure, and ligand-receptor interactions}~\cite{meyer2003}. Another example is given by the Polycyclic Aromatic Hydrocarbons (PAHs) as they  play a significant role in astrophysics and astrobiology.  PAHs (Fig.~\ref{fig:pah}) are thought to be among the most abundant and widespread organic molecules in the universe. They are identified in space via their unique infrared emission spectra~\cite{sandford2013infrared} and can form in the extreme conditions of space. Studying them can potentially contribute to our understanding of the formation of life's essential building blocks. 

\begin{figure}[t]
    \centering
    \includegraphics[scale= 0.69]{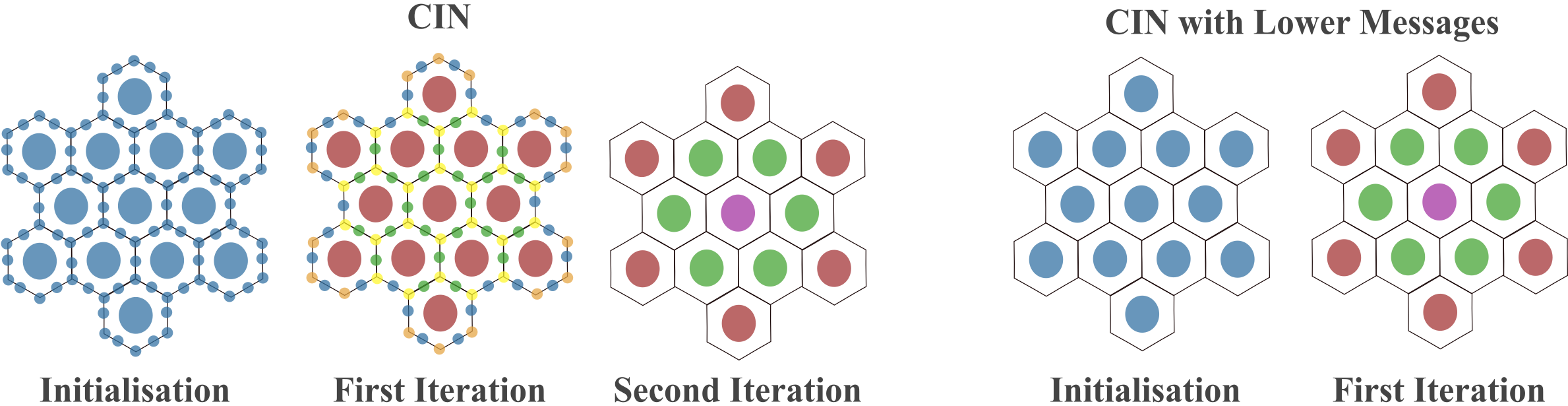}
    \caption{In molecular graphs featuring regions with a high concentration of rings, incorporating lower messages into cellular isomorphism networks expedites the convergence of the 2-cell colours.}
    \label{fig:ring coloring}
\end{figure}

\paragraph{On the convergence speed of Cellular Isomorphism Networks} 
Cellular Isomorphism Networks (CINs)~\cite{bodnarcwnet} are known to be powerful architectures able to model higher-order signals using a hierarchical message-passing procedure on cell complexes. Analysing the colouring procedure of CINs, the edges must first get the messages coming from the upper neighbourhood and only at the next iteration they can refine the colour of the rings (Fig.~\ref{fig:ring coloring} (left)). Although this colouring refinement procedure holds the same expressive power (\cite{bodnarcwnet}, Thm. 7), is it possible to reach a {\em faster convergence} by including messages coming from the lower neighbourhood of the cells. This allows for a direct interaction between the rings of the complex which removes the bottleneck caused by edges waiting for upper messages before updating ring colours (Fig.~\ref{fig:ring coloring} (right)).

\section{Enhancing Topological Message Passing}\label{sec:etmp}

\begin{wrapfigure}[13]{r}{0.225\textwidth}
    \begin{subfigure}[t!]{1.0\linewidth}
        \centering
        \vspace{+15pt}
        \includegraphics[width=1.0\textwidth]{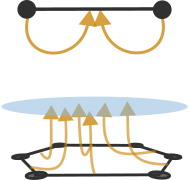}
    \end{subfigure}
     \caption{Boundary messages received by an edge (top) and a ring (bottom).}
    \label{fig:bm}
\end{wrapfigure}
\
\

In this section, we will describe our enhanced topological message-passing scheme that broadens the exchange of information within the cell complex. In particular, our enhancement consists of the inclusion of lower messages in the scheme proposed in~\cite{bodnarcwnet}. As we will show later in the section, including lower messages will let the information flow within a broader neighbourhood of the complex, enabling group interaction via the messages exchanged between the rings that are lower adjacent and escaping potential bottlenecks~\cite{alon2020bottleneck} via messages between lower adjacent edges. 

\subsection{Boundary Messages}

These are the messages that each cell $\sigma$ receives from its boundary elements $\tau \in \mathcal{B}(\sigma)$. We denote the information coming from the boundary of $\sigma$ as $m_\mathcal{B}(\sigma)$ and consists in a permutation invariant aggregation that takes as input all the {\em boundary messages} $M_{\mathcal{B}}$ between the feature vector $h_\sigma$ and all the feature vectors of its boundary elements, $h_\tau$ as in Fig.\ref{fig:bm}. Formally:

\begin{equation}\label{eq:boundary_msg}
    m_{\mathcal{B}}^{l+1}(\sigma) = \underset{\tau \in \mathcal{B}(\sigma)}{\text{AGG}}\Big(M_{\mathcal{B}}\big(h_{\sigma}^{l}, h_{\tau}^{l}\big)\Big). \nonumber
\end{equation}

This operation lifts the information from lower cells to higher-order ones, facilitating effective bottom-up communication across the complex. Leveraging the theory developed in~\cite{xu2018powerful} for graphs and later on in~\cite{bodnarcwnet} for regular cell complexes, to maximize the representational power of the underlying network, the function $m_{\mathcal{B}}$ is implemented using a Multi-Layer Perceptron (MLP) with 2 layers. 

\subsection{Upper Messages}
\begin{wrapfigure}[20]{r}{0.225\textwidth}
    \begin{subfigure}[t!]{1.0\linewidth}
        \centering
        \vspace{+10pt}
        \includegraphics[width=1.0\textwidth]{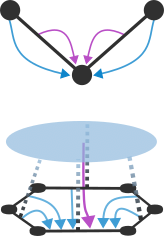}
    \end{subfigure}
     \caption{Upper messages are sent to a node (top) and to an edge (bottom). Co-boundary messages are shown in purple.}
    \label{fig:um}
\end{wrapfigure}
\
\

These are the messages that each cell $\sigma$ receives from its upper neighbouring cells $\tau \in \mathcal{\mathcal{N}^{{}^{\uparrow}}}(\sigma)$ (i.e., the blue arrows in Fig.\ref{fig:um}) and from common co-boundary cells $\delta \in \mathcal{C}o(\sigma, \tau)$ (i.e., the purple arrows in Fig.\ref{fig:um}). We denote the information coming from the upper neighbourhood of $\sigma$ and the common co-boundary elements as $m_{\mathcal{\mathcal{N}^{{}^{\uparrow}}}}$. It consists in a permutation invariant aggregation that takes as input all the {\em upper messages} $M_{\mathcal{\mathcal{N}^{{}^{\uparrow}}}}$ between the feature vector $h_\sigma$, all the feature vectors in its upper neighbourhood $h_\tau$ and all the cells in the common co-boundary neighbourhood, $h_\delta$. Formally:

\begin{equation}\label{eq:upp_msg}
    m_{\mathcal{\mathcal{N}^{{}^{\uparrow}}}}^{l+1}(\sigma) = \underset{\stackrel{\tau \in \mathcal{N}^{{}^{\uparrow}}(\sigma)}{{\delta \in \mathcal{C}o(\sigma, \tau)}}}{\text{AGG}}\Big(M_{\mathcal{N}^{{}^{\uparrow}}}\big(h_{\sigma}^{l}, h_{\tau}^{l}, h_{\delta}^{l}\big)\Big). \nonumber
\end{equation}

This operation will let the information flow within a  {\em narrow} neighbourhood of $\sigma$, ensuring consistency and coherence with respect to the underlying topology of the complex. We implement the function $m_{\mathcal{\mathcal{N}^{{}^{\uparrow}}}}$ using a 2-Layer MLP and $M_{{\mathcal{N}^{{}^{\uparrow}}}}$ is represented as a single dense layer followed by a point-wise non-linearity.

\subsection{Lower Messages}
\begin{wrapfigure}[15]{r}{0.225\textwidth}
    \begin{subfigure}[t!]{1.0\linewidth}
        \centering
        \vspace{-20pt}
        \includegraphics[width=1.0\textwidth]{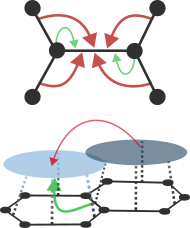}
    \end{subfigure}
     \caption{Lower messages are sent to an edge (top) and to a ring (bottom). Boundary messages are shown in green.}
    \label{fig:lm}
\end{wrapfigure}
\
\

These are the messages that each cell $\sigma$ receives from its lower neighbouring cells $\tau \in \mathcal{\mathcal{N}^{{}^{\downarrow}}}(\sigma)$ (i.e., the red arrows in Fig.\ref{fig:lm}) and from common boundary cells $\delta \in \mathcal{B}(\sigma, \tau)$ (i.e., the green arrows in Fig.\ref{fig:lm}). We denote a function that aggregates the information coming from the upper neighbourhood of $\sigma$ and the common co-boundary elements as $m_{\mathcal{\mathcal{N}^{{}^{\downarrow}}}}$. It consists in a permutation invariant aggregation that takes as input all the {\em lower messages} $M_{\mathcal{\mathcal{N}^{{}^{\downarrow}}}}$ between the feature vector $h_\sigma$, all the feature vectors in its lower neighbourhood $h_\tau$ and all the cells in the common boundary neighbourhood, $h_\delta$. Formally:

\begin{equation}\label{eq:lower_msg}
    m_{\mathcal{\mathcal{N}^{{}^{\downarrow}}}}^{l+1}(\sigma) = \underset{\stackrel{\tau \in \mathcal{N}^{{}^{\downarrow}}(\sigma)}{{\delta \in \mathcal{B}(\sigma, \tau)}}}{\text{AGG}}\Big(M_{\mathcal{N}^{{}^{\downarrow}}}\big(h_{\sigma}^{l}, h_{\tau}^{l}, h_{\delta}^{l}\big)\Big). \nonumber
\end{equation}

As pictorially shown in Fig.\ref{fig:lm} (top), this operation would help a {\em broader} diffusion of the information between edges that are not necessarily part of a ring. Also, it will let the rings of the complex communicate directly (Fig.\ref{fig:lm} (bottom)). Similarly to the upper messages, we implement $m_{\mathcal{\mathcal{N}^{{}^{\downarrow}}}}$ using an MLP with 2 layers. The function $M_{{\mathcal{N}^{{}^{\downarrow}}}}$ is implemented using a single dense layer followed by a point-wise non-linearity.

\subsection{Update and Readout}
Update and Readout operations are performed in line with~\cite{bodnarcwnet}. The exception is that in our case, the update function receives additional information provided by the messages that a cell $\sigma$ receives from its lower neighbourhood:

\begin{equation}\label{eq:update}
    h_{\sigma}^{l+1} = U\Bigl( h_\sigma^l, m_{\mathcal{B}}^{l}(\sigma), m_{\mathcal{\mathcal{N}^{{}^{\uparrow}}}}^{l+1}(\sigma),  m_{\mathcal{\mathcal{N}^{{}^{\downarrow}}}}^{l+1} (\sigma) \Bigr). 
\end{equation}
We represent the update function $U$ using a single fully connected layer followed by a point-wise non-linearity that uses a different set of parameters for each layer of the model and for each dimension of the complex.

After $L$ layers, we compute the representation of the complex $\mathcal{C}$ as: 

\begin{equation}\label{eq:readout}
    h_{\mathcal{C}} = R\Bigl( \{\{\{h_\sigma^L\}\}\}_{dim(\sigma)=0}^{2} \Bigr), 
\end{equation}

where $\{\{h_\sigma^L\}\}$ is the multi-set of cell's features at layer $L$. In practice, the representation of the complex is computed in two stages: first, for each dimension of the complex, we compute the representation of the cells at dimension $k$ by applying a mean or sum readout operation. This results in one representation for the vertices $h_\mathcal{V}$, one for the edges $h_\mathcal{E}$ and one for the rings $h_\mathcal{R}$. Then, we compute a representation for the complex $\mathcal{C}$ as: $h_{\mathcal{C}} = \mathrm{MLP}_{R,\mathcal{V}} \big( h_\mathcal{V} \big) + \mathrm{MLP}_{R,\mathcal{E}} \big( h_\mathcal{E} \big) + \mathrm{MLP}_{R,\mathcal{R}} \big( h_\mathcal{R} \big)$, where each $\mathrm{MLP}_{R,\cdot}$ is implemented as a single fully-connected layer followed by a non-linearity. Finally, $h_{\mathcal{C}}$ is forwarded to a final dense layer to obtain the predictions.

A neural architecture that updates the cell's representation using the message passing scheme defined in Eq.~\ref{eq:update} and obtains complex-wise representations as in Eq.~\ref{eq:readout} takes the name of {\em Enhanced Cell Isomorphism Network} (CIN++). The expressive power of CIN++ can then be directly inferred from the expressivity results reported in \cite{bodnarcwnet}.

\begin{theorem}
\label{thm:cin++express}
Let $\mathcal{F} : \mathcal{C} \rightarrow \mathbb{R}^d$ be a CIN++ network. With a sufficient number of layers and injective neighbourhood aggregators $\mathcal{F}$ is able to map any pair of
complexes $(\mathcal{C}_1, \mathcal{C}_2)$ in an embedding space that the Cellular Weisfeiler-Lehman (CWL) test is able to tell if $\mathcal{C}_1$ and  $\mathcal{C}_2$ are non-isomorphic.
\end{theorem}
\section{Experiments}
\begin{table}[t]
\vspace{-18pt}
\caption{Performance results on ZINC benchmark. We use gold \tikzcircle[gold,fill=gold]{2pt}, silver \tikzcircle[silver,fill=silver]{2pt}, and bronze \tikzcircle[bronze,fill=bronze]{2pt} colors to indicate the best performance.} 
  \vspace{2px}
  \label{tab:zinc-500k}
  \small
  \centering
  \resizebox{0.98\textwidth}{!}{ \renewcommand{\arraystretch}{1.0}
    \begin{tabular}{clcccc}
    \toprule
    \multirow{2}{*}{Method} & \multirow{2}{*}{Model} & \multirow{2}{*}{Time (s)} & \multirow{2}{*}{Params} & \multicolumn{2}{c}{Test MAE}\\
    & & & & ZINC-Subset & ZINC-Full \\ \midrule
    
    \multirow{7}{*}{MPNNs}
    & GIN~\citep{xu2019powerful} & 8.05 & 509,549 & 0.526$\pm$0.051 &  0.088$\pm$0.002 \\
    & GraphSAGE~\citep{hamilton2017inductive} & 6.02 & 505,341 & 0.398$\pm$0.002 & 0.126$\pm$0.003 \\
    & GAT~\citep{velivckovic2018graph} & 8.28 & 531,345 & 0.384$\pm$0.007 & 0.111$\pm$0.002 \\
    & GCN~\citep{kipf2017semisupervised} & 5.85 & 505,079 & 0.367$\pm$0.011 & 0.113$\pm$0.002 \\

    & MoNet~\citep{monti2017geometric} & 7.19 & 504,013 & 0.292$\pm$0.006 & 0.090$\pm$0.002 \\
    & \textls[-25]{GatedGCN-PE\citep{bresson2017residual}} & 10.74 & 505,011 & 0.214$\pm$0.006 & - \\
    & MPNN(sum)~\citep{gilmer2017neural} & - &  480,805 & 0.145$\pm$0.007 & - \\
    & PNA~\citep{corso2020principal} & - & 387,155 & 0.142$\pm$0.010 & - \\ 
    \midrule
    \multirow{2}{*}{\begin{tabular}[c]{@{}c@{}}Higher-order\\GNNs\end{tabular}} & RingGNN~\citep{chen2019equivalence} & 178.03 & 527,283 &  0.353$\pm$0.019 & - \\
    & 3WLGNN~\citep{maron2019provably} & 179.35 & 507,603 &  0.303$\pm$0.068 & - \\
    \midrule
    Substructure GNNs & GSN~\citep{bouritsas2022improving} & - & $\sim$500k & 0.101$\pm$0.010 & - \\
    \midrule
    
    \multirow{5}{*}{\begin{tabular}[c]{@{}c@{}}Subgraph\\GNNs\end{tabular}}
    & NGNN~\citep{zhang2021nested} & - & $\sim$500k & 0.111$\pm$0.003 & 0.029$\pm$0.001 \\
    & DSS-GNN~\citep{bevilacqua2022equivariant} & - & 445,709 & 0.097$\pm$0.006 & - \\
    & GNN-AK~\citep{zhao2022stars} & - & $\sim$500k & 0.105$\pm$0.010 & - \\
    & GNN-AK+~\citep{zhao2022stars} & - & $\sim$500k & 0.091$\pm$0.011 & - \\
    & SUN~\citep{frasca2022Understanding} & 15.04 & 526,489 & \third{0.083$\pm$0.003} & - \\ \midrule
    
    \multirow{4}{*}{\begin{tabular}[c]{@{}c@{}}Graph\\Transformers\end{tabular}}
    & GT~\citep{dwivedi2021generalization} & - & 588,929 & 0.226$\pm$0.014 & - \\
    & SAN~\citep{kreuzer2021rethinking} & - & 508,577 & 0.139$\pm$0.006 & - \\
    & Graphormer~\citep{ying2021transformers} & 12.26 & 489,321 & 0.122$\pm$0.006 & 0.052$\pm$0.005 \\ 
    & URPE~\citep{luo2022your} & 12.40 & 491,737 & 0.086$\pm$0.007 & \third{0.028$\pm$0.002} \\
    \midrule
    GD-WL & Graphormer-GD~\citep{zhang2023rethinking}   & 12.52 & 502,793 & ~~\second{0.081$\pm$0.009}& ~~\first{0.025$\pm$0.004} \\
    \midrule 

    \multirow{2}{*}{\begin{tabular}[c]{@{}c@{}}Topological NNs\end{tabular}} & CIN-Small~\citep{bodnar2021cellular} & - & $\sim$100k & 0.094$\pm$0.004 & 0.044$\pm$0.003 \\ 
     & CIN++ (ours) & 8.29 & 501,967 & ~~\first{0.077$\pm$0.004} & ~~\second{0.027$\pm$0.007} \\\bottomrule
    \end{tabular}
    }
    \vspace{-10pt}
\end{table}

In this section, we validate the properties of the proposed message-passing scheme in different real-world scenarios involving graph-structured data. We focus our experiments on a large-scale molecular benchmark (ZINC)~\cite{dwivedi2020benchmarking} and a long-range graph benchmark (Peptides)~\cite{dwivedi2022long}. Unless otherwise specified, in each Multi-Layer Perceptron, we apply Batch Normalization~\citep{BN} between the linear transformations and ReLU activations as used Adam~\cite{kingma2014adam}  with a starting learning rate of 0.001 that is halved whenever the validation loss reaches a plateau after a patience value we set to 20. We used an early stopping criterion that terminates the training when the learning rate reaches a threshold. Unless stated otherwise, we employ $1e^{-5}$ as the early stopping threshold.

\subsection{Large-Scale Molecular Benchmarks} 
We evaluate topological message passing on a large-scale molecular benchmark from the {\em ZINC} database~\cite{ZINCdataset}. The benchmark is composed of two datasets: {\em ZINC-Full} (consisting of 250K molecular graphs) and {\em ZINC-Subset} (an extract of 12k graphs from ZINC-Full) from~\cite{dwivedi2020benchmarking}. In these experiments, we used the same experimental setup of~\cite{bodnarcwnet} with the exception that we used 3 layers with a hidden dimension of 64. This restricts the parameter budget of our model to have nearly 500K parameters. We follow the training and evaluation procedures in~\cite{dwivedi2020benchmarking}. All results are illustrated in Tab.~\ref{tab:zinc-500k}. {\em Without any use of feature augmentation} such as positional encoding, our model exhibits particularly strong performance on these benchmarks: it attains state-of-the-art results by a significant margin on {\em ZINC-Subset},  outperforming other models by a significant margin and is on par with the best baselines for {\em ZINC-Full}.

\subsection{Long-Range Graph Benchmarks}

\begin{table}[t]
    \caption{Performance results for \pepfunc (graph classification) and \pepstruct (graph regression). Best scores are highlighted using gold \tikzcircle[gold,fill=gold]{2pt}, silver \tikzcircle[silver,fill=silver]{2pt}, and bronze \tikzcircle[bronze,fill=bronze]{2pt} colors.}
    \label{tab:experiments_peptides}
    \begin{adjustwidth}{-2.5 cm}{-2.5 cm}\centering
    \scalebox{0.9}{
    \setlength\tabcolsep{4pt} 
    \begin{tabular}{l c c c c}\toprule
    \multirow{2}{*}{\textbf{Model}} & \multicolumn{2}{c}{\pepfunc} & \multicolumn{2}{c}{\pepstruct} \\\cmidrule(lr){2-3}\cmidrule(lr){4-5}
    &\textbf{Train AP} &\textbf{Test AP $\uparrow$} &\textbf{Train MAE} &\textbf{Test MAE $\downarrow$} \\\midrule
    MLP & 0.4217$\pm$0.0049 & 0.4060$\pm$0.0021 & 0.4273$\pm$0.0011 & 0.4351$\pm$0.0008 \\
    GCN &0.8840$\pm$0.0131 &0.5930$\pm$0.0023 &0.2939$\pm$0.0055 &0.3496$\pm$0.0013 \\
    GCNII  & 0.7271$\pm$0.0278 & 0.5543$\pm$0.0078 & 0.2957$\pm$0.0025 & 0.3471$\pm$0.0010\\
    GINE  &0.7682$\pm$0.0154 &0.5498$\pm$0.0079 &0.3116$\pm$0.0047 &0.3547$\pm$0.0045 \\
    GatedGCN  &0.8695$\pm$0.0402 &0.5864$\pm$0.0077 &0.2761$\pm$0.0032 &0.3420$\pm$0.0013 \\
    GatedGCN+RWSE  &0.9131$\pm$0.0321 &0.6069$\pm$0.0035 &0.2578$\pm$0.0116 &0.3357$\pm$0.0006 \\ \midrule
    Transformer+LapPE  &0.8438$\pm$0.0263 & 0.6326$\pm$0.0126 &0.2403$\pm$0.0066 &\second{0.2529$\pm$0.0016} \\
    SAN+LapPE  &0.8217$\pm$0.0280 &\third{0.6384$\pm$0.0121} &0.2822$\pm$0.0108 &0.2683$\pm$0.0043 \\
    SAN+RWSE  &0.8612$\pm$0.0219 &\second{0.6439$\pm$0.0075} &0.2680$\pm$0.0038 & \third{0.2545$\pm$0.0012} \\ \midrule
    CIN++ (ours)  &0.8943$\pm$0.0226&\first{0.6569$\pm$0.0117} &0.229$\pm$0.0079 &\first{0.2523$\pm$0.0013} \\
    \bottomrule
    \end{tabular}
    }
    \vspace{-10pt}
    \end{adjustwidth}
\end{table}

To test the effectiveness of enhanced topological message passing for discovering long-range interactions we evaluate our method on a long-range molecular benchmark~\cite{dwivedi2022long}. The datasets used from the benchmark are derived from 15,535 peptides that compose the SATPdb database~\cite{singh2016satpdb}. For this benchmark, we evaluate our method against the tasks of peptides structure prediction (\pepstruct) and peptides function prediction (\pepfunc). For both datasets, we did not employ any feature augmentation such as positional encoding. We ensured that the parameter budget was constrained to 500K. We repeat the training with 4 different seeds and report the mean of the test AP and MAEs at the time of early stopping. For \pepstruct, we used a cellular lifting map that considers all the induced cycles of dimension up to 8 as rings. We used 3 layers with 64 as a hidden dimension, a batch size of 128 and a sum aggregation to obtain complex-level embeddings. For \pepfunc, we attach 2 cells to all the induced cycles of dimension up to 6. We used 4 layers with an embedding dimension of 50, and a batch size of 64. A dropout~\cite{srivastava2014dropout} with a probability of 0.15 is inserted.  With respect to the other benchmarks, we set the starting learning rate of $4e-4$, a weight decay of $5e^{-5}$. The final readout is performed with a mean aggregation. As shown in Tab.~\ref{tab:experiments_peptides} we achieve very high performance on these tasks even without any use of feature augmentation.

\section{Conclusions, Related Works and Future Developments}

\paragraph{Broader Impacts} 
This work provides evidence of how the enhanced topological message-passing scheme proposed in this work allows the integration of local and global information within a cell complex in the context of computational chemistry. In particular, our model captures complex dependencies and long-range interactions more effectively. We foresee the proposed work having a broad impact within the field of computational chemistry, as our scheme offers a robust and versatile approach to predict meaningful properties of chemical systems by accurately modelling complex dependencies and capturing long-range and group interactions.


\paragraph{Related Works}
In light of Topological Deep Learning being an emerging research area that has been introduced quite recently~\cite{hajijtopological}, numerous pioneering works appeared in this field.  
In~\cite{bodnar2021weisfeiler} the authors proposed a Simplicial Weisfeiler-Lehman (SWL) colouring procedure for distinguishing non-isomorphic simplicial complexes and a provably powerful message passing scheme based on SWL, that generalise Graph Isomorphism Networks~\cite{xu2018powerful}. This was later refined in~\cite{bodnarcwnet}, where the authors introduced CW Networks (CWNs), a hierarchical message-passing on cell complexes proven to be strictly more powerful than the WL test and not less powerful than the 3-WL test. In~\cite{hajij2020cell}, the authors provide a general message-passing mechanism over cell complexes however, they do not study the expressive power of the proposed scheme, nor its complexity. Furthermore, they did not experimentally validate its performance. The works in~\cite{sheaf2022, suk2022surfing} introduced Neural Sheaf Diffusion Models, neural architectures that learn a sheaf structure on graphs to improve learning performance on  transductive tasks in heterophilic graphs. Meanwhile, attentional schemes appeared in topological deep learning in the context of simplicial complexes~\cite{giusti22, anonymous2022SAT}, cellular complexes~\cite{giusti2022cell}, sheaves~\cite{barbero2022sheaf} and combinatorial complexes~\cite{hajij2022}. For a more detailed examination of the architectures developed in the field of topological deep learning, we refer the reader to the survey of Papillon~\cite{papillon2023architectures}. Recent works considered also rings within the message passing scheme by means of Junction Trees (JT)~\cite{Fey2020_himp} and by augmenting node features with information about cycles~\cite{bouritsas2020improving}. However, it is easy to see that these schemes have a different design than the one provided in this work.

\paragraph{Limitations} 
While our work demonstrates that topological message passing effectively models complex dependencies and long-range interactions in chemical systems, we acknowledge that the complexity of the proposed method inherently increases due to the cellular lifting maps and the additional messages sent throughout the complex. We mitigate this computational overhead by mapping all the graphs present in the datasets into cell complexes in a pre-processing stage and storing them for later use. Additionally, the overhead of our message-passing scheme is mitigated by the fact that the operations within the same layer are naturally decoupled. Efficient network implementations make it possible to update the representation of a cell $\sigma$ in a concurrent execution~\cite{Besta22parallel}, amortizing the cost to be proportional to the largest neighbourhood of $\sigma$.

\paragraph{Conclusions}
Our study has presented an innovative approach to neural networks operating on graph-structured data. Current state-of-the-art models do not naturally account for a principled way to model group interactions. We addressed this by introducing an enhancement of the Topological Message Passing scheme developed in~\cite{bodnarcwnet}. The newly proposed Topological Message Passing scheme, named CIN++, enables a direct interaction within high-order structures of the underlying cell complex, by letting messages flow within its lower neighbourhood without sacrificing the model's expressivity. By allowing the exchange of messages between higher-order structures, we significantly enhance the model's capacity to capture multi-way relationships in the data. We have demonstrated that the ability to model long-range and group interactions is critical for capturing real-world chemistry-related problems. In particular, the natural affinity of cellular complexes for representing higher-dimensional structures and topological features will provide a more detailed understanding of complex chemical systems compared to traditional models.

\bibliography{ref}


\newpage
\appendix

\section{Glossary}
\begin{table}[!htb]
\begin{center}
\caption{Summary of Notations. The first section, {\em Structural Elements}, details the notation for fundamental mathematical and topological constructs such as graphs, regular cell complexes, and their associated components. The second section, {\em Functional Elements}, delineates the notation for functional aspects, including feature vectors, information exchange, and various functional operations.}
\label{tab:multicol}
\begin{tabular}{ll}
\hline
\multicolumn{2}{Sc}{\textbf{Structual Elements}}\\
\toprule
$\;\mathcal{G} = (\mathcal{V},\mathcal{E})$ & A graph, $\mathcal{V}$ and $\mathcal{E}$ are respectively the sets of vertices and edges.\\\vspace{2pt}
$\mathcal{C} = (\mathcal{V}, \mathcal{E}, \mathcal{R})$ & A regular cell complex, $\mathcal{R}$ is the set of rings.\\\vspace{2pt}
$v_i$ & An element of $\mathcal{V}$. \\\vspace{2pt}
$e_i  = (v_i, v_j) $ & An element of $\mathcal{E}$. \\\vspace{2pt}
$r_i = (e_{i_1}, \ldots, e_{i_{\vert r_i \vert}} )$ & An element of $\mathcal{R}$. $\vert r_i \vert$ is the size of the $i$-th ring. \\\vspace{2pt}
$\mathcal{B}({\sigma})$ & Boundary of $\sigma$.  \\\vspace{2pt}
$\mathcal{C}o({\sigma})$ &  Co-boundary of $\sigma$. \\ \vspace{2pt}
$\mathcal{N}^{{}^{\uparrow}}({\sigma})$ & Upper neighbourhood of $\sigma$.  \\\vspace{2pt}
$\mathcal{N}^{{}^{\downarrow}}({\sigma})$ & Lower neighbourhood of $\sigma$.  \\\vspace{2pt}
$\sigma \face \tau$ & Boundary relationship (i.e. $\sigma \in \mathcal{B}(\tau)$ ). \\\vspace{2pt}
$\mathcal{B}({\sigma, \tau})$ & Boundary elements in common between $\sigma$ and $\tau$.  \\\vspace{2pt}
$\mathcal{C}o({\sigma, \tau})$ & Co-boundary elements in common between $\sigma$ and $\tau$. \\
\bottomrule
\multicolumn{2}{Sc}{\textbf{Functional Elements}}\\
\toprule
$\;h_{\sigma}^l$ & Feature vector of the cell $\sigma$ at layer $l$.\\\vspace{2pt}
$m_{\mathcal{B}}^l(\sigma)$ & Information that $\sigma$ receives from  cells $\tau \in \mathcal{B}(\sigma)$ at layer $l$.\\\vspace{2pt}
$m_{\mathcal{N}^{{}^{\uparrow}}}^l(\sigma)$ & Information that $\sigma$ receives from cells $\tau \in \mathcal{N}^{{}^{\uparrow}}(\sigma), \delta \in \mathcal{C}o(\sigma, \tau)$ at layer $l$. \\\vspace{2pt}
$m_{\mathcal{N}^{{}^{\downarrow}}}^l(\sigma)$ & Information that $\sigma$ receives from cells $\tau \in \mathcal{N}^{{}^{\downarrow}}(\sigma), \delta \in \mathcal{B}(\sigma, \tau)$ at layer $l$. \\\vspace{2pt}
$M_{\mathcal{B}}$ & Boundary message function between $\sigma$ and  $\tau \in \mathcal{B}(\sigma)$.\\\vspace{2pt}
$M_{\mathcal{N}^{{}^{\uparrow}}}$ & Upper message function between $\sigma$, $\tau \in \mathcal{N}^{{}^{\uparrow}}(\sigma)$ and $\delta \in \mathcal{C}o(\sigma, \tau)$. \\\vspace{2pt}
$M_{\mathcal{N}^{{}^{\downarrow}}}$ & Lower message function between $\sigma$, $\tau \in \mathcal{N}^{{}^{\downarrow}}(\sigma)$  and $\delta \in \mathcal{B}(\sigma, \tau)$. \\\vspace{2pt}
$AGG$ & Permutation equivariant aggregation function. \\\vspace{2pt}
$U$ & Update function.\\\vspace{2pt}
$R$ & Readout function. \\
\bottomrule
\end{tabular}
\end{center}
\end{table}

\section{Expressive Power}\label{app:exp}
In this section we analyse the expressive power of enhanced topological message passing. Two complexes $\mathcal{C}_1$ and $\mathcal{C}_2$ are said to be {\em isomorphic} (written $\mathcal{C}_1 \simeq \mathcal{C}_2$) if there exists a bijection $\varphi:\mathcal{P}_{\mathcal{C}_1} \to \mathcal{P}_{\mathcal{C}_2}$ such that $\sigma \in \mathcal{C}_1 \iff \varphi(\sigma) \in \mathcal{C}_2$~\cite{bodnar2021topological, bodnarcwnet}. Also, we say that a cell coloring $c$ {\em refines} a cell coloring $d$, written $c \sqsubseteq d$, if $c(\sigma) = c(\tau) \implies d(\sigma) = d(\tau)$ for every $\sigma,\tau \in \mathcal{C}$. 
Two colorings are {\em equivalent} if $c \sqsubseteq d$ and $d \sqsubseteq c$, and we write $c \equiv d$~\cite{morris2019weisfeiler}.

\begin{proof}[\textbf{Proof of Theorem~\ref{thm:cin++express}}]
Let $c^l$ be the colouring of CWL~\cite{bodnarcwnet} at iteration $l$ and $h^l$ the colouring (i.e., the multi-set of features) provided by a CIN++ network at layer $l$ as in Section \ref{sec:etmp}.

To show that CIN++ inherits all the properties of Cellular Isomorphism Networks~\cite{bodnarcwnet} we must show that our scheme produces a colouring of the complex that satisfies Lemma 26 of~\cite{bodnarcwnet}. 

To show $c^t \sqsubseteq h^t$ by induction, let us assume $h^l = h^L$ for all $l > L$, where $L$ is the number of the network's layers.
Let also $\sigma, \tau$ be two arbitrary cells with $c^{l+1}_\sigma = c^{l+1}_\tau$. Then, $c^l_\sigma = c^l_\tau$, $c^l_\gB(\sigma) = c^l_\gB(\tau)$, $c^l_\uparrow(\sigma) = c^l_\uparrow(\tau)$ and $c^l_\downarrow(\sigma) = c^l_\downarrow(\tau)$. By the induction hypothesis, $h^l_\sigma = h^l_\tau$, $h^l_\gB(\sigma) = h^l_\gB(\tau)$, $h^l_\uparrow(\sigma) = h^l_\uparrow(\tau)$ and $h^l_\downarrow(\sigma) = h^l_\downarrow(\tau)$. 

If $l+1 > L$, then $h^{l+1}_\sigma = h^{l}_\sigma = h^l_\tau = h^{l+1}_\tau$. Otherwise, $h^{l+1}$ is given by the update function in Equation \ref{eq:update}. Given that the inputs passed to these functions are equal for $\sigma$ and $\tau$, $h^{l+1}_\sigma = h^{l+1}_\tau$.

For showing  $h^l \sqsubseteq c^l$, let us suppose the aggregation from Equation \ref{eq:update} is injective and the model is equipped with a sufficient number of layers such that the convergence of the colouring is guaranteed. Let $\sigma, \tau$ be two cells with $h^{l+1}_\sigma = h^{l+1}_\tau$. Then, since the local aggregation is injective $h^l_\sigma = h^l_\tau$, $h^l_\gB(\sigma) = h^l_\gB(\tau)$, $h^l_\uparrow(\sigma) = h^l_\uparrow(\tau)$ and $h^l_\downarrow(\sigma) = h^l_\downarrow(\tau)$. By the induction hypothesis, $c^l_\sigma = c^l_\tau$, $c^l_\gB(\sigma) = c^l_\gB(\tau)$,  $c^l_\uparrow(\sigma) = c^l_\uparrow(\tau)$ and   $c^l_\downarrow(\sigma) = c^l_\downarrow(\tau)$ which implies that $c^{l+1}_\sigma = c^{l+1}_\tau$. 

Given that we have shown $c^t \sqsubseteq h^t$ and $h^l \sqsubseteq c^l$, we can conclude that $h^l \equiv c^l$.
\end{proof}

As a result, we have that CIN++ inherits all the properties of Cellular Isomorphism Networks, in accordance with Lemma 26 from~\cite{bodnarcwnet}.
\section{Complexity Analysis}
In this section we will characterise the complexity of a CIN++ layer in terms of number of messages and number of parameters involved for a two-dimensional regular cell complex $\mathcal{C}$ whose number of cells is denoted with $\vert \mathcal{C} \vert$. Let us assume $\sigma \in \mathcal{C}$ be an arbitrary cell of $\mathcal{C}$  such that dim$(\sigma)$ is either  0, 1 or 2 and $\vert \mathcal{B}(\sigma) \vert \leq c_1$ and  $\vert \mathcal{C}o(\sigma) \vert \leq c_2$, for some constants $c_1$ and $c_2$. Let also $h_{\sigma}^{l} \in \mathbb{R}^d$ be the features of the cell $\sigma$ at layer $l$.

\paragraph{Boundary Messages} The boundary message function is implemented as:

\begin{equation}\label{eq:bm_impl}
    m_{\gB}^l(\sigma) :=  \mathrm{MLP}^{l}_{\gB}\big( (1+\eps_{\gB}) h_{\sigma}^{l} + \sum_{\tau \in \gB(\sigma)} h_{\tau}^{l} \big), \nonumber
\end{equation}

where $\mathrm{MLP}^{l}_{\gB}$ has 2 fully-connected layers. Considering that 0-cells (vertices)  do not have boundary elements, 1-cells (edges) have only two boundary elements and the maximum ring size of $\mathcal{C}$ is bounded by a small constant~\cite{bodnarcwnet}, we have that the number of boundary messages scales with $\mathcal{O}(\vert \mathcal{C} \vert)$.  The number of parameter involved in this operation is $\mathcal{O}(d^2)$, provided by the outer Multi-Layer Perceptron (MLP). In this work, we did not employ any parameter sharing across the dimensions of the complex (i.e., we used a distinct perceptron for each layer of the network and for each dimension of the complex).

\paragraph{Upper Messages} We implement the upper message function as:

\begin{equation}\label{eq:um_impl}              
    m_{{\mathcal{N}^{{}^{\uparrow}}}}^l(\sigma) := \mathrm{MLP}^{l}_{{\mathcal{N}^{{}^{\uparrow}}}}\big( (1+\eps_{{\mathcal{N}^{{}^{\uparrow}}}}) h_{\sigma}^{l} + \sum_{\stackrel{\tau \in \mathcal{N}^{{}^{\uparrow}}(\sigma)}{{\delta \in \mathcal{C}o(\sigma, \tau)}}} \mathrm{MLP}^{l}_{M_{\uparrow}} \big( h_{\tau}^{l} \parallel h_{\delta}^l \big) \big), \nonumber
\end{equation}

In this context, $\mathrm{MLP}^{l}_{M{\uparrow}}$ denotes a single-layer fully-connected network, complemented by a point-wise non-linearity, while $\mathrm{MLP}^{l}_{{\mathcal{N}^{{}^{\uparrow}}}}$ represents a two-layer Multilayer Perceptron (MLP). As indicated in~\cite{bodnarcwnet}, the amount of upper messages that a cell $\tau \in \mathcal{B}(\sigma)$ exchanges with its adjacent cells is given by $2 \cdot \binom{\left| \mathcal{B}(\sigma) \right|}{2}$. Considering the assumption that the boundary of the cells is bounded by a fixed constant, the total number of messages correlates linearly with the magnitude of the complex, that is, the number of cells in $\mathcal{C}$. The amount of learnable parameters is also on the order of $\mathcal{O}(d^2)$, a consequence of the two MLPs utilized in the message function.

\paragraph{Lower Messages} The implementation of the lower message function is as follows:

\begin{equation}\label{eq:lm_impl}              
    m_{{\mathcal{N}^{{}^{\downarrow}}}}^l(\sigma) := \mathrm{MLP}^{l}_{{\mathcal{N}^{{}^{\downarrow}}}}\big( (1+\eps_{{\mathcal{N}^{{}^{\downarrow}}}}) h_{\sigma}^{l} + \sum_{\stackrel{\tau \in \mathcal{N}^{{}^{\downarrow}}(\sigma)}{{\delta \in \mathcal{B}(\sigma, \tau)}}} \mathrm{MLP}^{l}_{M_{\downarrow}} \big( h_{\tau}^{l} \parallel h_{\delta}^l \big) \big). \nonumber
\end{equation}

As for the upper messages, $\mathrm{MLP}^{l}_{M{\downarrow}}$ denotes a single-layer fully-connected network, succeeded by a point-wise non-linearity, while $\mathrm{MLP}^{l}_{{\mathcal{N}^{{}^{\downarrow}}}}$ represents a two-layer Multilayer Perceptron (MLP). The amount of lower messages that a cell $\tau \in \mathcal{C}o(\sigma)$ exchanges with its neighbours is given by $2 \cdot \binom{\left| \mathcal{C}o(\sigma) \right|}{2}$. Since we assumed also that the cells have a number of co-boundary neighbours that is bounded by a fixed constant, the total number of messages scales linearly with the number of cells in the complex. The two MLPs involved in the message function induces an amount of learnable parameters on the order of $\mathcal{O}(d^2)$.

\section{A Categorical Interpretation: Sheaves}
Our method can be seen as a particular case of a message passing scheme over a cellular sheaf. 
Let $\mathcal{C}$ be a regular cell complex. A cellular sheaf is a mathematical object that attaches data spaces to the
cells of $\mathcal{C}$ together with relations that specify when assignments to these data
spaces are consistent.

\begin{definition}[Cellular Sheaf]\cite{hansen2019toward}\label{def:sheaf} A \style{cellular sheaf} of vector spaces on a regular cell complex $\mathcal{C}$ is an
assignment of a vector space $\Fc(\sigma)$ to each cell $\sigma$ of $\mathcal{C}$
together with a linear transformation $\Fc_{\sigma\face\tau}\colon \Fc(\sigma)
\to \Fc(\tau)$ for each incident cell pair $\sigma\,\face\,\tau$. These must
satisfy both an identity relation $\Fc_{\sigma\face\sigma}=\id$ and the
composition condition:
\[
     \rho\,\face\,\sigma\,\face\,\tau
  ~~\Rightarrow~~
  \Fc_{\rho\face\tau} = \Fc_{\sigma\face\tau}\circ\Fc_{\rho\face\sigma}.
\]
\end{definition}

It is also natural to consider a dual construction to a
cellular sheaf to preserves stalk data but reverses
the direction of the face poset, and with it, the restriction maps.
\begin{definition}[Cellular Cosheaf]\cite{hansen2019toward}\label{def:cosheaf}
A cellular cosheaf of vector spaces on a regular cell complex $\mathcal{C}$ is an
assignment of a vector space $\Fc(\sigma)$ to each cell $\sigma$ of  $\mathcal{C}$
together with linear maps $\Fc^{\text{op}}_{\sigma\face\tau}\colon \Fc(\tau) \to
\Fc(\sigma)$ for each incident cell pair $\sigma\,\face\,\tau$ which satisfies
the identity ($\Fc^{\text{op}}_{\sigma\face\sigma}=\id$) and composition condition:
\[
     \rho\,\face\,\sigma\,\face\,\tau
  ~~\Rightarrow~~
  \Fc^{\text{op}}_{\rho\face\tau} = \Fc^{\text{op}}_{\rho\face\sigma}\circ\Fc^{\text{op}}_{\sigma\face\tau} .
\]
\end{definition}

The vector space $\Fc(\sigma)$ is called the {\em stalk} of $\Fc$ at $\sigma$ and will encode the features supported over $\sigma$. The maps $\Fc_{\sigma\face\tau}$ and $\Fc^{\text{op}}_{\sigma\face\tau}$ are called the {\em restriction maps} and will provide a principled way to respectively move features from lower dimensional cells to higher dimensional ones and vice-versa. 


From a categorical perspective, a cellular sheaf is a functor $\Fc : \mathcal{P}_{\mathcal{C}} \rightarrow \mathbf{Vect}_{\mathbb{R}}$ that maps the indexing set $\mathcal{P}_{\mathcal{C}}$ to the category of vector spaces over $\mathbb{R}$ while a cellular cosheaf is a functor $ \Fc^{\text{op}} : \mathcal{P}_{\mathcal{C}}^{\text{op}} \to
\mathbf{Vect}_{\mathbb{R}}$ such that, if we consider a two dimensional regular cell complex $\mathcal{C}$, a sheaf $(\Fc, \mathbb{R})$ and its dual cosheaf $(\Fc^{\text{op}}, \mathbb{R})$ on  $\mathcal{C}$, the following diagram commutes:

\begin{wrapfigure}[5]{r}{0.4\textwidth}
    \begin{subfigure}[t!]{1.0\linewidth}
        \centering
        \includegraphics[width=1.0\textwidth]{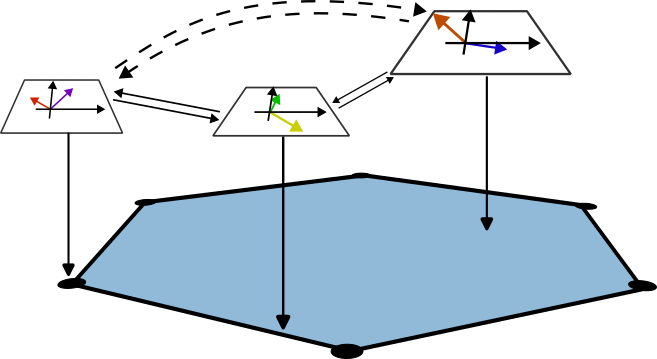}
    \end{subfigure}
     \caption{Pictorial example of a sheaf and cosheaf of vector spaces structure on a ring of a regular cell complex $\mathcal{C}$.}
    \label{fig:sheaf}
    \vspace{-60pt}
\end{wrapfigure}

\
\

\[\begin{tikzcd}[sep=2.25em]
	{\Fc(v)} &&& {\Fc(e)} \\
	\\
	\\
	&&& {\Fc(r)}
	\arrow["{\Fc_{v\face e}}", from=1-1, to=1-4]
	\arrow["{\Fc_{e \face r}}", from=1-4, to=4-4]
	\arrow["{\Fc_{v \face r}}"', shift right=1, curve={height=30pt}, dashed, from=1-1, to=4-4]
	\arrow["{\Fc_{v \face r}^{\text{op}}}"', shift right=1, curve={height=-30pt}, dashed, from=4-4, to=1-1]
	\arrow["{\Fc_{e \face r}^{\text{op}}}", shift left=2, from=4-4, to=1-4]
	\arrow["{\Fc_{v \face e}^{\text{op}}}", shift left=2, from=1-4, to=1-1]
\end{tikzcd}\]

\vspace{5pt}
In the given commutative diagram, the arrow is dashed to indicate that the morphism (map) it represents is not explicitly defined in the diagram, but rather it is implied by the other morphisms. In this case, the dashed arrow is used to show the existence of a unique morphism that makes the diagram commute. This relationship is important in the context of cellular sheaves, where these morphisms represent restrictions on different cells and their overlaps. The dashed arrows shows that there's a unique way to go from $\Fc(v)$ to $\Fc(r)$  and back that is consistent with the other restrictions, even if it's not directly defined in the diagram.

\section{Additional Experimental Details}

\subsection{Computational Resources and Code Assets} 

In all experiments we used NVIDIA\textsuperscript{\textregistered} Tesla V100 GPUs with 5,120 CUDA cores and 32GB GPU memory on a personal computing platform with an Intel\textsuperscript{\textregistered} Xeon\textsuperscript{\textregistered} Gold 5218 CPU @ 2.30GHz CPU using Ubuntu 18.04.6 LTS.

The model has been implemented in PyTorch~\citep{NEURIPS2019_9015} by building on top of CW Networks library\footnote{\url{https://github.com/twitter-research/cwn/}}~\citep{bodnarcwnet} and PyTorch Geometric library\footnote{\url{https://github.com/pyg-team/pytorch_geometric/}}~\citep{fey2019fast}. High-performance lifting operations use the graph-tool\footnote{\url{https://graph-tool.skewed.de/}} Python library and are parallelised via Joblib\footnote{\url{https://joblib.readthedocs.io/en/latest/}}.
PyTorch, NumPy, SciPy and Joblib are made available under the BSD license, Matplotlib under the PSF license, graph-tool under the GNU LGPL v3 license. CW Networks and PyTorch Geometric are made available under the MIT license.

\subsection{Large Scale Molecular Benchmarks}

\begin{table}[!t]
    \centering
    \begin{minipage}[t]{0.80\textwidth}
        \centering
         \caption{ZINC-Subset (MAE), ZINC-Full (MAE) and Mol-HIV.}
        \label{tab:mol_dataset}
           \resizebox{\columnwidth}{!}{
          \begin{tabular}{l ccc}
            \toprule
            \multirow{2}{*}{Model} & 
            
            ZINC-Subset  &
            ZINC-Full & 
            MOLHIV\\
            
            &
            (MAE $\downarrow$) &
            (MAE $\downarrow$)&
            (ROC-AUC $\uparrow$) \\
            \midrule
            
            GCN \citep{kipf2017graph} & 
            0.469$\pm$0.002 &
            N/A &
            76.06$\pm$0.97 \\

            GAT \citep{velivckovic2017graph} & 
            0.463$\pm$0.002 &
            N/A &
            N/A \\
            
            GatedGCN \citep{bresson2017residual} &
            0.363$\pm$0.009 &
            N/A &
            N/A \\

            GIN \citep{GIN}  & 
            0.252$\pm$0.014 &
            0.088$\pm$0.002 &
            77.07$\pm$1.49 \\
            
            PNA \citep{Corso2020_PNA} & 
            0.188$\pm$0.004 &
            N/A &
            79.05$\pm$1.32 \\
            
            DGN \citep{beaini2020directional} & 
            0.168$\pm$0.003 &
            N/A &
            79.70$\pm$0.97 \\
            
            HIMP \citep{Fey2020_himp} &
            0.151$\pm$0.006 &
            0.036$\pm$0.002 &
            78.80$\pm$0.82\\

            GSN \citep{bouritsas2020improving} & 
            0.108$\pm$0.018 &
            N/A &
            77.99$\pm$1.00 \\
            
            \midrule
            
            CIN-small \cite{bodnarcwnet} & 
            0.094$\pm$0.004 &
            \third{0.044$\pm$0.003} &
            \third{80.55$\pm$1.04} \\
            
            CIN \cite{bodnarcwnet} & 
            \second{0.079$\pm$0.006} &
            \second{0.022$\pm$0.002} &
            \first{80.94$\pm$0.57} \\

            \midrule

            \textbf{CIN++-small} (ours) & 
            \third{0.091$\pm$0.003} &
            0.044$\pm$0.004 &
            80.26$\pm$1.02 \\
            
            \textbf{CIN++} (ours) & 
            \first{0.074$\pm$0.004} &
            \first{0.021$\pm$0.001} &
            \second{80.63$\pm$0.94} \\
            
            \bottomrule
          \end{tabular}%
          }
    \end{minipage}
\end{table}

\paragraph{ZINC}
The number of nodes (or atoms) in the graphs ranges from 3 to 132, with an average size of approximately 24 nodes. The majority of the graphs have between 10 and 30 nodes. The average degree in the graphs is approximately 2 and the average diameter of the graphs is approximately 12.4 nodes (or atoms) and the maximum diameter was 62 nodes. Regarding the edges (or bonds), the average number of edges in the graphs is approximately 50 composed of 98\% by single bonds, while the remaining 2\% are aromatic bonds. These are two graph regression task datasets for drug-constrained solubility prediction which have been built on top of the ZINC database provided by the Irwin and Shoichet Laboratories in the Department of Pharmaceutical Chemistry at the University of California, San Francisco (UCSF)~\cite{ZINCdataset}. Each graph represents a molecule, where the features over the nodes specify which atom it represents while edge features specify the type of chemical bond between two atoms. Graph-level targets correspond to the penalised water-octanol partition coefficient -- logP, an important metric in drug design that depends on chemical structures and molecular properties and characterizes the drug-likeness of a molecule~\cite{gomez2018automatic}.

\paragraph{MOLHIV}

In our study, we further validate our model using the \texttt{ogbg-molhiv} molecular dataset from the Open Graph Benchmark~\citep{hu2020open}. Eeach graph is a representation of a molecule, where the nodes stand for atoms and the edges symbolize chemical bonds. The node features, which are 9-dimensional, include the atomic number, chirality, and other atom-specific attributes such as formal charge and ring inclusion. The edge features, which are 3-dimensional, incorporate the bond type, bond stereochemistry, and an additional feature that indicates the presence of a conjugated bond. The statistics of the graphs in the dataset are similar to the ones discussed for the ZINC benchmark. The objective is to ascertain the ability of compounds to inhibit HIV replication. We consider a maximum ring size of 6 nodes as 2-cells. Our model's architecture and hyperparameter settings mirror those referenced in previous studies~\cite{bodnarcwnet, Fey2020_himp}. In Tab.~\ref{tab:mol_dataset}, we present the average test ROC-AUC metrics at the epoch of optimal validation performance across 10 random weight initializations. For this dataset, we experience lower performance than

\subsection{Long-Range Molecular Benchmarks}
In both tasks of this benchmark, each graph corresponds to a peptide molecule~\cite{dwivedi2022long}. Peptides, in the realm of biology, are depicted as compact polymers of amino acids, which are covalently bonded through peptide linkages formed between the carboxyl group of one amino acid and the amino group of another. These molecules execute a diverse spectrum of functions in living organisms, serving as signaling molecules, protective agents of the immune system, structural constituents, transporters, enzymes, and even as a nutritional source. In biological systems, peptides manifest as short polymers of amino acids connected via peptide bonds - a linkage established between the carboxyl group of one amino acid and the amino group of another. These entities perform a comprehensive array of functions in living organisms. For instance, they function as signaling molecules, safeguarding agents of the immune system, structural components, transporters, and enzymes. Moreover, they also serve as a source of nutrition~\cite{singh2016satpdb}. Since each amino acid is composed of many heavy atoms, the molecular graph of a peptide is much larger than that of a small drug-like molecule. The long-range molecular benchmark proposes two datasets for \peptides property prediction where the graphs are derived such that the nodes correspond to the heavy (non-hydrogen) atoms of the peptides while the edges represent the bonds that join them. The peptides datasets have about 5 times larger diameter ($\approx$ 57) and 6 times more atoms than the molecular graphs present in the {\em ZINC} benchmark ($\approx$ 151) and the average node degree is 2.04. The average shortest path is 20.89. The requirements for long-range interactions and sensitivity to the graph's global properties are met through the three-dimensional structural dependencies intrinsic to the peptide chains combined with a substantial raise in number of nodes in the graphs.

\subsection{TUDataset}

\begin{table}[!t]
    \centering
    \caption{TUDatasets. The first part shows the performance of graph kernel methods. The second assess graph neural networks while the third part is for topological neural networks. We mark the models with highest performance using gold \tikzcircle[gold,fill=gold]{2pt}, silver \tikzcircle[silver,fill=silver]{2pt}, and bronze \tikzcircle[bronze,fill=bronze]{2pt} colors.}
    \label{tab:tud}
    \resizebox{\linewidth}{!}{%
    \begin{tabular}{l  lllll}
        \toprule
        Model & 
        MUTAG &
        PTC\_MR &
        PROTEINS &
        NCI1 &
        NCI109\\
        \midrule       
        RWK \citep{gartner2003graph} & 
         79.2$\pm$2.1 & 
         55.9$\pm$0.3 & 
         59.6$\pm$0.1 & 
         $>$3 days & 
         N/A \\
        
        GK ($k=3$) \citep{shervashidze2009efficient} &
        81.4$\pm$1.7 & 
        55.7$\pm$0.5 & 
        71.4$\pm$0.3 &  
        62.5$\pm$0.3 & 
        62.4$\pm$0.3  \\

        PK \citep{neumann2016propagation} & 
         76.0$\pm$2.7& 
         59.5$\pm$2.4 & 
         73.7$\pm$0.7 & 
         82.5$\pm$0.5 & 
         N/A  \\

        WL kernel \citep{shervashidze2011weisfeiler} &
          90.4$\pm$5.7 & 
          59.9$\pm$4.3 & 
          75.0$\pm$3.1 & 
          \first{86.0$\pm$1.8} & 
          N/A \\

        \midrule
         
        DCNN \citep{DCNN_2016} & 
        N/A&  
        N/A & 
        61.3$\pm$1.6 & 
        56.6$\pm$1.0 & 
        N/A  \\

        DGCNN \citep{zhang2018end} & 
        85.8$\pm$1.8 & 
        58.6$\pm$2.5 & 
        75.5$\pm$0.9 & 
        74.4$\pm$0.5 & 
        N/A \\
        
        IGN \citep{maron2019invariant} &
        83.9$\pm$13.0 & 
        58.5$\pm$6.9 & 
        76.6$\pm$5.5 & 
        74.3$\pm$2.7 & 
        72.8$\pm$1.5 \\
        
        GIN \citep{GIN} & 
        89.4$\pm$5.6 & 
        64.6$\pm$7.0 &  
        76.2$\pm$2.8 & 
        82.7$\pm$1.7 &  
        N/A \\

        PPGNs \citep{maron2019provably} &
        90.6$\pm$8.7 & 
        66.2$\pm$6.6 & 
        \third{77.2$\pm$4.7} &  
        83.2$\pm$1.1 & 
        82.2$\pm$1.4  \\

        Natural GN \citep{de2020natural} &
        89.4$\pm$1.6 & 
        66.8$\pm$1.7 & 
        71.7$\pm$1.0 & 
        82.4$\pm$1.3 &  
        N/A \\

        GSN \citep{bouritsas2020improving} &
        92.2 $\pm$ 7.5 & 
        68.2 $\pm$ 7.2 & 
        76.6 $\pm$ 5.0 & 
        83.5 $\pm$ 2.0 &  
        N/A \\ 
       
        \midrule
        
        SIN \citep{bodnar2021weisfeiler} & 
        N/A  & 
        N/A &  
        76.4 $\pm$ 3.3 & 
        82.7 $\pm$ 2.1 & 
        N/A \\ 

        CIN \citep{bodnarcwnet} & 
        \third{92.7 $\pm$ 6.1} & 
        \third{68.2 $\pm$ 5.6} & 
        77.0 $\pm$ 4.3 & 
        83.6 $\pm$ 1.4 & 
        \second{84.0 $\pm$ 1.6}  \\ 
        
        CAN \citep{giusti2022cell} & 
        \second{94.1 $\pm$ 4.8} & 
        \second{72.8 $\pm$ 8.3} & 
        \second{78.2 $\pm$ 2.0} &  
        \third{84.5 $\pm$ 1.6}  & 
        \third{83.6 $\pm$ 1.2}  \\ 

        \midrule
       {{\bf CIN++} (Ours)} & 
        \first{94.4 $\pm$ 3.7} & 
        \first{73.2 $\pm$ 6.4} & 
        \first{80.5 $\pm$ 3.9} & 
        \second{85.3 $\pm$ 1.2} & 
        \first{84.5 $\pm$ 2.4}  \\ 
        \bottomrule

    \end{tabular}
    }
\end{table}

The \texttt{TUDataset}~\cite{TUDataset} is a rich repository of graph-based datasets, serving as a benchmark for learning tasks on graph-structured data. Specifically, we took the ones within the domain of small molecules and bioinformatics. The \texttt{MUTAG} dataset, for instance, comprises nitroaromatic compounds, where the task is to predict their mutagenicity on Salmonella typhimurium~\cite{debnath1991structure}. The dataset is structured as graphs, with vertices representing atoms labeled by atom type and edges representing bonds between the corresponding atoms, consisting of 188 samples of chemical compounds with 7 discrete node labels. Another  dataset used is \texttt{PTC}, a collection of 344 chemical compounds, each represented as a graph, with the goal to report carcinogenicity for rodents, and 19 node labels for each node~\cite{toivonen2003statistical}.

The \texttt{NCI1} and \texttt{NCI109} and dataset, from the cheminformatics domain, represents each chemical compound as a graph, where vertices and edges respectively symbolize atoms and bonds between atoms. The dataset pertains to anti-cancer screens with chemicals evaluated for their effectiveness against cell lung cancer~\cite{wale2008comparison}. Each vertex label denotes the corresponding atom type, encoded via a one-hot-encoding scheme into a binary vector. The \texttt{PROTEINS} dataset is utilized in the field of bioinformatics for protein function prediction~\cite{borgwardt2005protein}. The task is to predict functional class membership of enzymes and non-enzymes. We assess the performance of enhanced topological message passing scheme against graph kernel methods, graph neural networks as well as topological neural networks. In this assessment, we employed the same model configurations used in~\cite{bodnarcwnet}.  In Table~\ref{tab:tud} we report that the proposed scheme achieves state of the art results on four out of five different evaluations. The exception is for  \texttt{NCI1} where our method achieves the second place after WL kernel~\cite{shervashidze2011weisfeiler}.

\end{document}